\documentclass[letterpaper, 10 pt, conference]{ieeeconf}  

\IEEEoverridecommandlockouts                              

\usepackage{graphicx} 
\usepackage[ruled, linesnumbered]{algorithm2e}
\usepackage{hyperref}
\usepackage{multirow}
\usepackage{comment}
\usepackage{mathptmx}
\usepackage{subfloat}
\usepackage{subfig}
\usepackage{diagbox}
\usepackage{epstopdf}
\usepackage{pifont}
\usepackage{amsmath}
\usepackage{multirow}
\usepackage{url}
\usepackage{verbatim}
\usepackage{footmisc}
\usepackage{color}
\usepackage{tabularx}
\usepackage{float}
\usepackage{booktabs}
\usepackage{makecell}
\usepackage{bm}
\usepackage{hyperref}
\usepackage[utf8x]{inputenc}
\usepackage{graphicx}
\title{\LARGE \bf
	Intelligent LiDAR Navigation: Leveraging External Information and Semantic Maps with LLM as Copilot*
}

\author{Fujing Xie$^{1}$, Jiajie Zhang$^{1}$, S\"{o}ren Schwertfeger$^{1}$
	\thanks{*This work has been partially funded by the Shanghai Frontiers Science Center of Human-centered Artificial Intelligence.
		The experiments of this work were supported by the core facility Platform of Computer Science and Communication, SIST, ShanghaiTech University}
	\thanks{
		$^{1}$The authors are with the Key Laboratory of Intelligent Perception and Human-Machine Collaboration -- ShanghaiTech University, Ministry of Education, China.{ \{xiefj,zhangjj2023,soerensch\}@shanghaitech.edu.cn}}%
}

\begin{document}
	
	\captionsetup{font={footnotesize}}
	
	\maketitle
	\thispagestyle{empty}
	\pagestyle{empty}

	\begin{abstract}
		
		Traditional robot navigation systems primarily utilize occupancy grid maps and laser-based sensing technologies, as demonstrated by the popular move\_base package in ROS. 
		Unlike robots, humans navigate not only through spatial awareness and physical distances but also by integrating external information, such as elevator maintenance updates from public notification boards and experiential knowledge, like the need for special access through certain doors.
		With the development of Large Language Models (LLMs), which possesses text understanding and intelligence close to human performance, there is now an opportunity to infuse robot navigation systems with a level of understanding akin to human cognition.
		In this study, we propose using osmAG (Area Graph in OpensStreetMap textual format), an innovative semantic topometric hierarchical map representation, to bridge the gap between the capabilities of ROS move\_base and the contextual understanding offered by LLMs.
		Our methodology employs LLMs as an actual copilot in robot navigation, enabling the integration of a broader range of informational inputs, while maintaining the robustness of traditional robotic navigation systems. 
		Our code, demo, map, and experiment results can be accessed at \href{https://github.com/xiexiexiaoxiexie/Intelligent-LiDAR-Navigation-LLM-as-Copilot}{https://github.com/xiexiexiaoxiexie/Intelligent-LiDAR-Navigation-LLM-as-Copilot}. 
		
	\end{abstract}

	\section{INTRODUCTION}
\begin{figure}[t]
		\centering
		\includegraphics[width=0.40\textwidth]{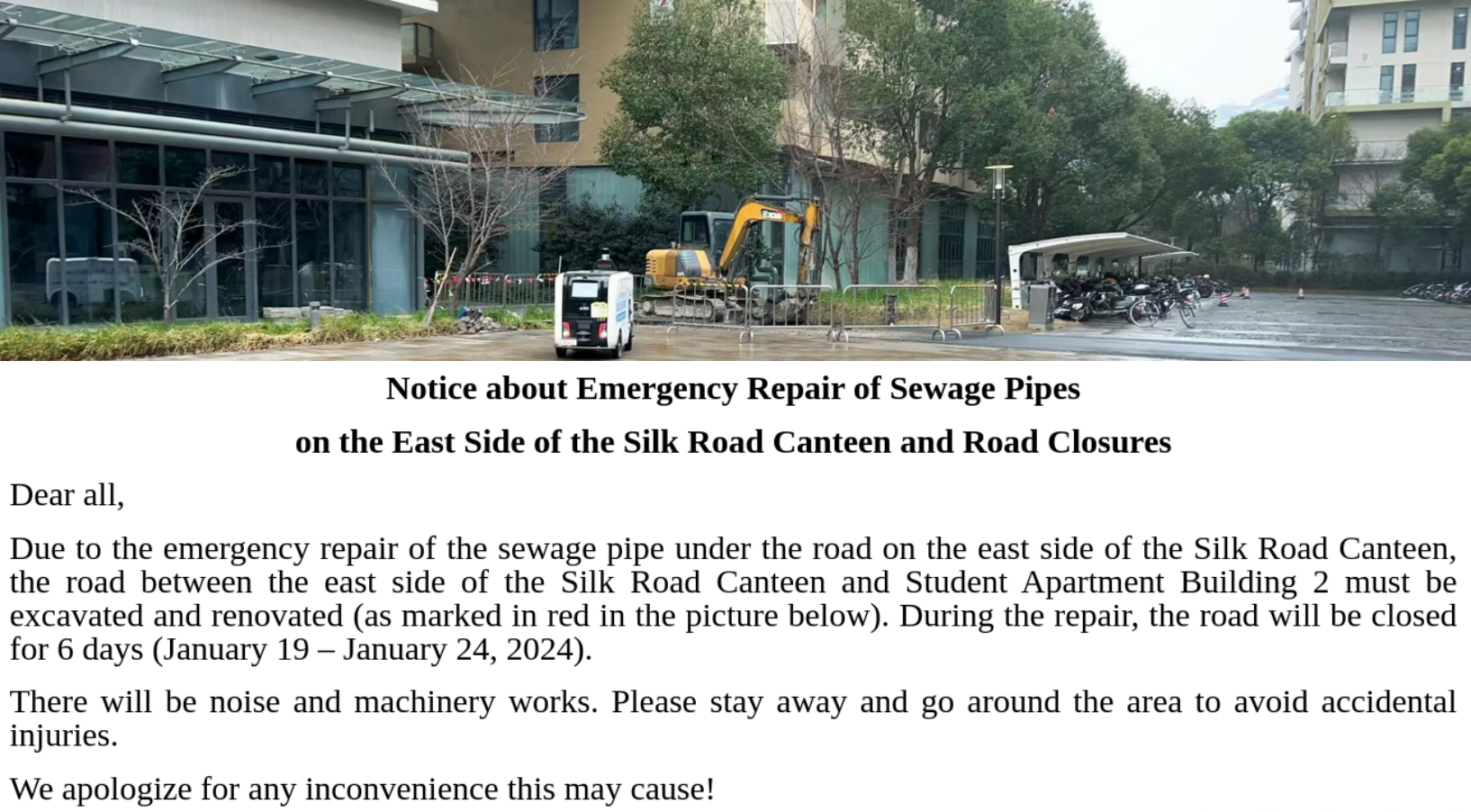}
		\caption{ The figure above depicts a real-life situation encountered by a 3rd-party delivery robot on our University campus, where it is blocked by an intersection closure. Below the e-mail sent by Office of General Services announcing this closure is shown. {\color{black}Reprinted from \cite{xie2024Empowering}.}}	
		\label{fig:email}
		\vspace{-8mm}
	\end{figure}

	In real-world scenarios, robots operating in large dynamic environments face significant challenges, {\color{black}as their surroundings can not always be sensed or updated instantly}. An illustrative, real-life example is a campus delivery robot encountering a path obstruction due to an unanticipated pipe repair closure, depicted in Fig. \ref{fig:email}. Despite prior notice on public websites, the robot's navigation algorithm remained unaware of this obstacle. With the capabilities of LLMs, which can process general text-based information, we now have the opportunity to integrate human-like contextual understanding into robotic systems. This way, in the above example, the robot could, by reading the website, from the beginning plan a path that avoids the construction areas, thus achieving a faster and shorter delivery route.

	In recent years, the integration of LLMs or VLMs (Vision-Language Models) into robot navigation has opened up diverse avenues for research and application, as highlighted in \cite{lin2023advances}. 
	Most of these approaches rely on real-time mapping rather than using an a-priori map, as seen in works like \cite{yu2023l3mvn,zhou2024navgpt,gadre2023cows}, often provide only the next immediate step, lacking long-term planning capabilities.
	However, without a prior map, LLMs are not naturally aware of a robot's specific environment. 
	For instance, if asked where to find a pair of scissors, an LLM might suggest searching in a kitchen. When provided with a map of a campus building that includes a robotics lab, the LLM can more accurately direct the robot to the lab.
	Therefore maps serve as a tool for proper grounding LLMs to the robot's environment is essential. 
	
	For the scenario of this paper we argue, that in robot navigation, we can utilize an existing map that represents the environment and has semantic information. This map, in our case in the osmAG format \cite{feng2023osmAG}, can be created by a Simultaneous Localization and Mapping (SLAM) approach or from CAD files. 
	Utilizing LiDAR sensors we can localize the robot in this map
	and employ A* for optimal path planning once the goal coordinates are known.
	In our paper we utilize two LLM modules for two distinct tasks: 1) For human-robot-interaction we need to interpret the given language command to identify the navigation goal.  For that the LLM  leverages the semantic information encoded in the map to find goal coordinates, e.g. by finding the right room for ``Please bring this item to the robotics training lab".  This module also utilized experience data to potentially adjust map costs. 2) We interpret external information to adjust map costs.


	
	\begin{figure*}[ht]
		\centering
		\includegraphics[width=0.88\textwidth]{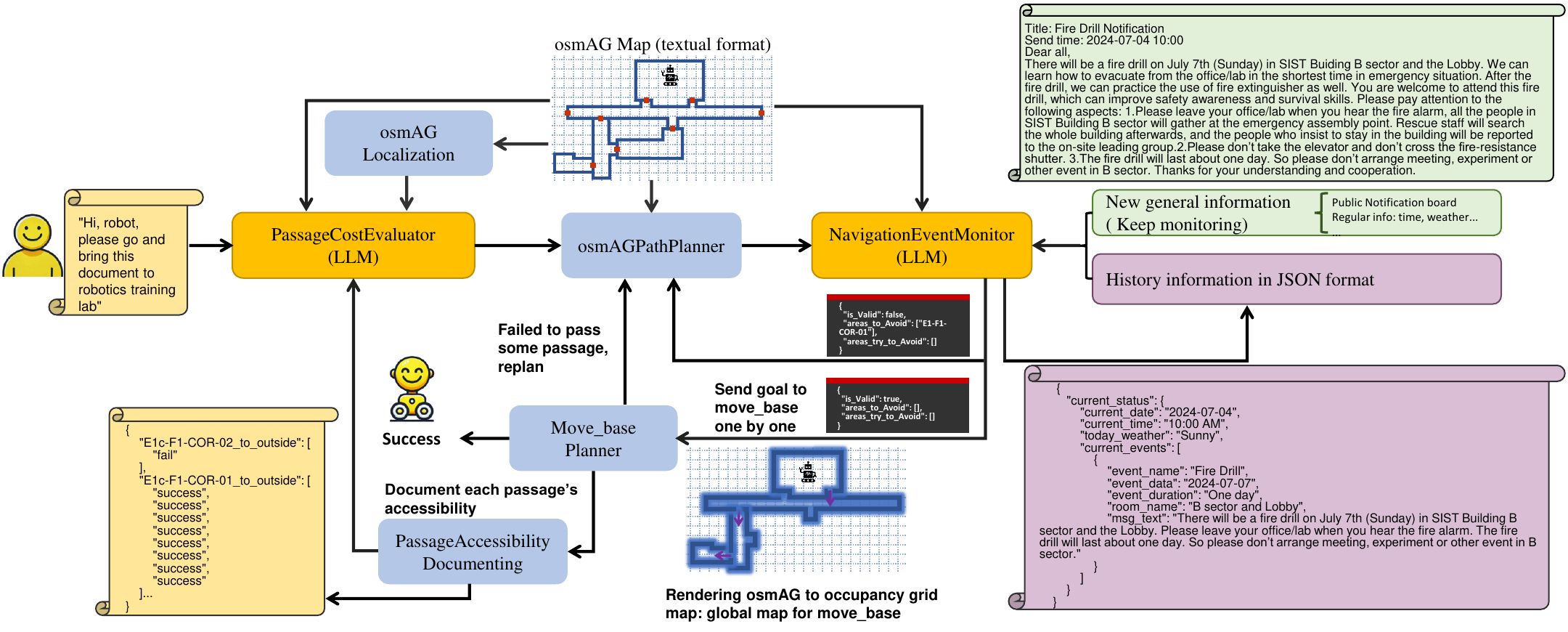}
		\caption{ Pipeline of our algorithm: The \textit{NavigationEventMonitor} keeps tracking events from public notifications and stores information that could impact navigation. When receiving a human instruction, the \textit{PassageCostEvaluator} identifies the destination and assess passage costs based on instructions, accessibility document, and osmAG. This data is used by \textit{osmAGPathPlanner} to plan a path, which is then sent to the \textit{NavigationEventMonitor} for approval. If approved, the path (a list of passages) is sent to move\_base to move the robot. If not, \textit{NavigationEventMonitor} suggests areas to avoid, prompting \textit{osmAGPathPlanner} to generate an alternative path until approval is granted.}	
		\label{fig:flow}
		\vspace{-7mm}
	\end{figure*}

	When interpreting external information, the LLM needs to match the description of locations (e.g. a specific elevator) with the right entity in the map.  This highlights the importance of developing a map representation that is interpretable by both the LLM and laser-based navigation systems, such as move\_base, to enable seamless integration of these technologies.

	The osmAG map we propose to use in this paper is a hierarchical, topometric semantic map that represents physical spaces as polygon based areas, such as rooms and corridors. It is compact and textual to ensure compatibility with LLMs for token efficiency and ease of human editing. In  \cite{xie2023robust}, we presented a robust and lifelong LiDAR-based localization method that utilize osmAG to find the robot pose by matching the sensor data to the permanent building structures (walls, doors). 
	The map is compatible with traditional robotic algorithms and minimizes update frequency by focusing on permanent structural information.

	
	In this paper, we explore the potential of LLMs serving as co-pilots that understand external information to assist existing laser-based robot navigation systems within large environments. For example, in our experiment, our robot operates in a campus building of approximately 6200 $m^2$ and 70+ rooms for one floor.
	Our work can be quickly employed to real robots with a laser based sensor that are already operating in reality and still benefit from the general knowledge LLM process.

%
	
	The pipeline of our approach is illustrated in Fig. \ref{fig:flow}. Our inputs include the osmAG map, human instructions, saved experience files from previous navigation tasks, and tracked events from external notifications. 
	We use real notifications sent by our university administration as the primary source of external information, along with general data such as current time and weather. However, the system isn’t limited to just this—humans can easily add other information or preferences into the prompt, allowing the system to adapt to specific needs or situations, e.g. by subscribing to email newsletters from the campus administration and social media channels. 
	
	Our experimental results demonstrate that by leveraging LLMs as a copilot, our approach augments traditional laser-based navigation with socially and contextually relevant information, such as public notifications, thereby enhancing robustness and adaptability in human-centric environments.
	
	Our contributions are as follows:
	\begin{itemize}
		\item [$\bullet$]Our system leverages LLMs to comprehend map data in osmAG format, combining this with historical data and external information sources like public notifications.
		\item [$\bullet$]
		We integrate LLMs into laser-based navigation systems using the same osmAG map, enabling straightforward implementation on real-world robots equipped with LiDARs and an osmAG map, leveraging their ability to plan optimal paths with algorithms like A*. 
		\item [$\bullet$] We make our code, simulation environments, and experimental results open-source, allowing others to test different LLMs or navigation techniques.
		
	\end{itemize}

	\section{Related Works}

	\subsection{LLM Guided Robotics}
	
	Recent research has integrated robots with LLMs to enhance human interaction and decision making. PaLM-E \cite{driess2023palm} combines real-world sensor data with language models for improved multimodal reasoning, while \cite{brohan2023rt} fine-tunes PaLM-E using robotic trajectory data to generate actions. Other works utilize LLMs for tasks like object rearrangement \cite{ding2023task}, creating semantic cost maps for motion planning \cite{sharma2022correcting}, encoding high-level instructions into actionable trajectories \cite{bucker2023latte}, and enabling robots to follow and generate navigation instructions \cite{wang2023lana}. Additionally, \cite{kannan2024smart} introduces a framework for embodied multi-robot task planning. All these approaches introduce the broad knowledge and reasoning capabilities of LLMs into robotics.
	\subsection{LLM Guided Navigation}
	Most LLM-guided navigation research focuses on using cameras as sensors because visual data makes extracting semantic information easier compared to LiDAR data. For instance, \cite{vemprala2024chatgpt} achieves basic navigation by asking an LLM to generate code that calls high-level functions like `turn\_left' to control robot movement. \cite{shah2023lm} navigates robots based on natural language instructions, such as ``After passing a white building, take a right next to a white truck." 
	\cite{chen2022think}, \cite{huang2023visual} and \cite{yokoyama2024vlfm} build maps using real time sensor data instead of utilizing existing maps.
	However, these approaches either rely on navigation commands without using proven algorithms like $A^*$, or they treat path planning as part of mission planning, leaving the LLM unaware of the specific path. In contrast, our paper aims to integrate LLMs directly into the path planning and navigation system, combining the speed and efficiency of $A^*$ with the broad knowledge and intelligence of LLMs for optimal navigation.

	\vspace{-0.5mm}
	\subsection{Semantic maps in Robotics}
	\vspace{-1mm}
	Armeni et al. \cite{armeni20193d} and Hughes et al. \cite{hughes2022hydra} introduced the Scene Graph, an RGB-D camera-based 3D environment representation that organizes spatial concepts into layered graphs, from detailed semantic meshes to buildings. Later works \cite{chen2024scene, rana2023sayplan, dai2024optimal, gu2024conceptgraphs} applied Scene Graphs to language-based localization and task planning. Unlike Scene Graphs, osmAG focuses on permanent structures, unaffected by occlusions and changes. Additionally, osmAG-based algorithms don't rely on visual data, aligning more closely with traditional navigation algorithms like move\_base.
	\vspace{-0.0mm}
	\section{Approach}
	\vspace{-0.0mm}
	
\begin{figure}[t]
		\centering
		\includegraphics[width=0.49\textwidth]{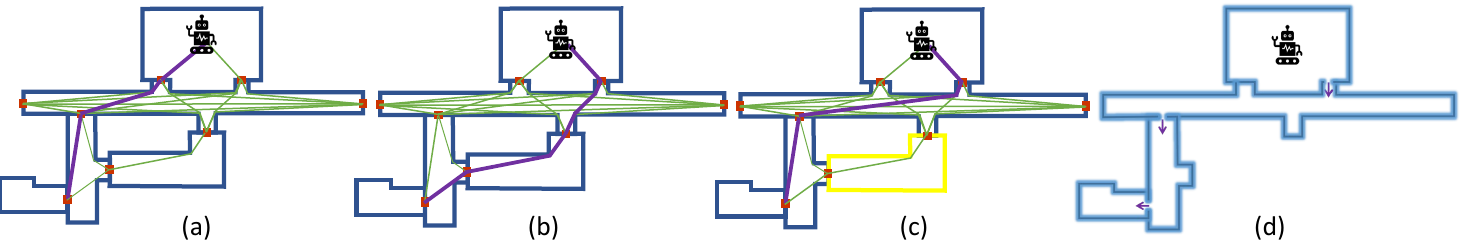}
		\caption{ \textit{osmAGPathPlanning} process: Blue polygons represent areas, red lines denote passages, green paths indicate driving distances between passage pairs within an area (edge cost in $AG$, it is map-specific and stored in a file), and purple paths show the current shortest path. (a) Shortest path based on distance alone; (b) \textit{PassageCostEvaluator} favors the right passage in the start area due to consistently open passages; (c) \textit{NavigationEventMonitor} advises avoiding the yellow-marked area, likely occupied due to an event; (d) Final path approved by the \textit{NavigationEventMonitor}, with each passage sent as a goal to move\_base. The occupancy grid map is rendered as the move\_base global map, with other passages closed to enforce path adherence.
		}	
		\label{fig:pipe_map}
		\vspace{-4mm}
	\end{figure}

	The osmAG map representation is comprehensively defined in \cite{feng2023osmAG}. As illustrated in Fig. \ref{fig:pipe_map} and the osmAG map used in our experiments (Fig. \ref{fig:case_path}), it represents physical spaces as polygon-based areas, such as rooms and corridors. Passages are represented as line segments of the area polygons, connecting neighboring areas (e.g., doors). A formal definition of the osmAG is provided in Section \ref{osmAG}.

	When navigating familiar buildings, humans rely on experiential knowledge—such as avoiding doors that require special access or choosing automatic doors when their hands are full—as well as external information like elevator maintenance notification.
	Inspired by this behavior, our system incorporates two key modules to enhance robot navigation. First, as illustrated in Fig. \ref{fig:flow}, our system employs a module called  \textit{PassageCostEvaluator} (described in \ref{Passage_cost}), which monitors whether the robot successfully passes through specific passages and records this information in a file. This file is then used to adjust cost of passages, enabling the \textit{osmAGPathPlanner} to generate more informed and efficient paths based on past navigation experiences.
	Second, the system employs the \textit{NavigationEventMonitor} (described in \ref{NavigationEventMonitor}) to continuously track external notifications. This module saves relevant information into a file, which is used during navigation to assess the validity of the path planned by the \textit{osmAGPathPlanner}. How the \textit{osmAGPathPlanner} plans a path is described in Section. \ref{osmAG Path Planner}, and the integration with ROS move\_base is in Section. \ref{move_base}.
	
	The use of two files and two LLM-related modules is motivated by the following considerations: When robots navigate large buildings, as shown in Fig. \ref{fig:pipe_map}, the path is determined by selecting which passages to traverse, as movement between areas requires determinate which passages to pass. 
	However, human communication about locations often refers to general areas rather than specific passages. For example, humans might indicate a fire drill in Sector B of a building, without specifying which doors will be affected. This difference requires to treat passages and areas differently.

	\begin{figure*}[t]
		\centering
		\includegraphics[width=0.88\textwidth]{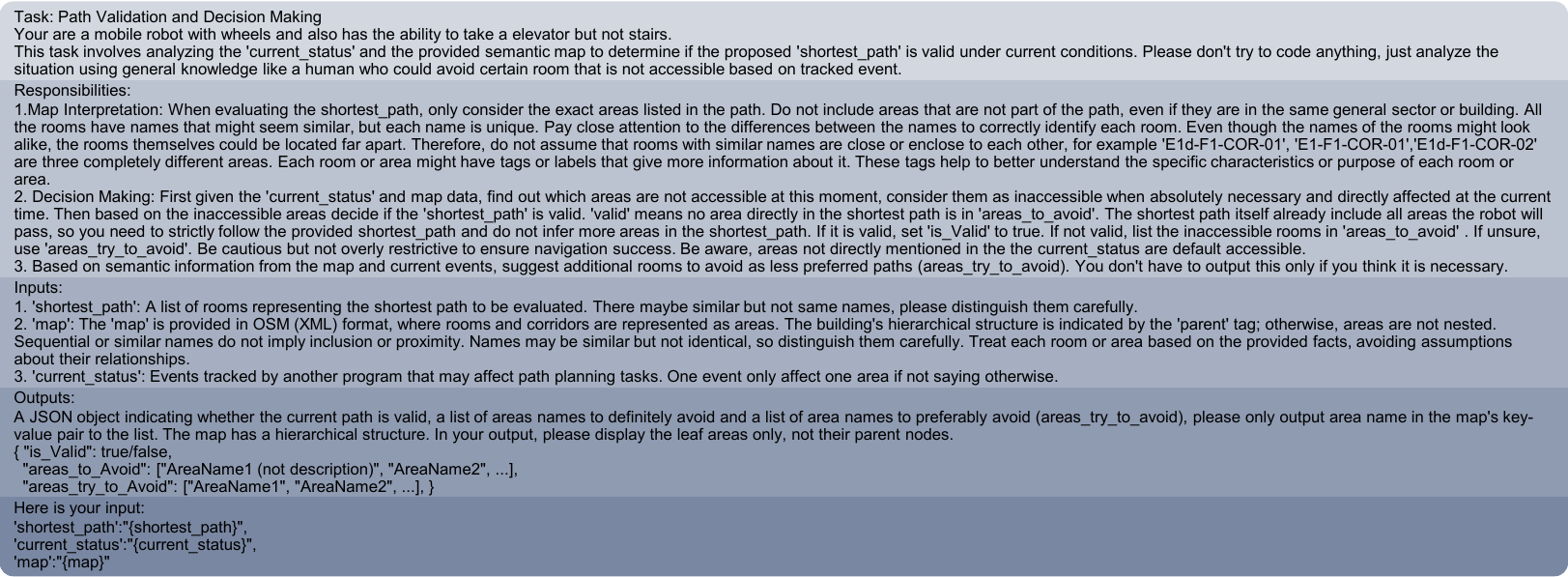}
		\caption{ The prompt for the \textit{NavigationEventMonitor} module instructs the LLM to assess the validity of a robot’s navigation path. The prompt begins with a high-level task description, followed by detailed instructions for the LLM. It then outlines the input parameters and defines the expected output format. Finally, the prompt provides the necessary data for evaluation, including the path being assessed, the tracked current status, and the semantic osmAG map.	}
		\label{fig:prompt}
		\vspace{-7mm}
	\end{figure*}
	
	\vspace{-0.5mm}

	\subsection{osmAG Map Representation}
	\vspace{-0.5mm}
	\label{osmAG}

	For our navigation task, the topology of the osmAG is defined as $AG = (V, E)$,  where $V$ represents vertices (passages) and $E$ represents edges (pairs of passages associated with the same area). 
	The cost attached to each edge $d_{ij}$ is the driving distance between a pair of passages.
	As illustrated by the green paths in Fig. \ref{fig:pipe_map}, for each $AG$, we use A* to pre-calculate the values of $d_{ij}$ based on the polygon of this area. We store these values in a file as they are map-specific and remain constant over time. 
	This file is subsequently used by another A* algorithm in \textit{osmAGPathPlanner} to compute the optimal sequence of passages from the robot position $P_{robot}$ to the goal area $A_{goal}$, denoted as $\mathbf{P}=( v_{1}, v_{2}, \ldots, v_{k})$, where $ v_{1}, v_{2}, \ldots, v_{k}$ are the intermediate vertices (passages).

	\vspace{-0.5mm}

	\subsection{Human Instruction Processing and Passage Cost Evaluation}
	\vspace{-0.5mm}
	\label{Passage_cost}
	As illustrated in Fig. \ref{fig:flow}, initially, the system receives a human instruction, such as `Hi, robot, please go and bring this document to the robotics training lab.' Using the osmAG provided in the prompt, the LLM identifies the destination area $A_{goal}$.
	In addition to the edge costs $d_{ij}$ described in Section \ref{osmAG}, our approach incorporates an additional cost $pc_{i}$ for traversing individual passages.
	This cost $pc_{i}$ is determined based on two factors: the type of passage (automatic or handle-operated doors) and the robot's prior experience with it. 
	Some mobile manipulation robots can open doors independently if their hands are free, while others depend on assistance from humans or other robots, or must find alternative routes. As a result, automatic doors may be assigned lower costs compared to doors with handles. Experience data is collected as detailed in Section \ref{sec:experience}. Certain doors, such as those frequently blocked by obstacles, are often difficult to open and cause repeated failures, leading the LLM to assign higher costs to these consistently problematic passages.
	
	The prompt to assess $pc_{i}$ follows the structure illustrated in Fig. \ref{fig:prompt}, includes an overall description, the LLM's responsibilities, and input/output formats, but without examples or chain-of-thought reasoning. For brevity, the specific prompt is omitted here but is available in our Git repository.
	The output of this module includes the identified destination area $A_{goal}$ and the $pc_{i}$ of all passages.

	\vspace{-1mm}
	\subsection{Navigation Event Tracking and Path Validation}
	\vspace{-1mm}
	\label{NavigationEventMonitor}
	As shown in Fig. \ref{fig:flow}, the \textit{NavigationEventMonitor} is a second LLM that has two main tasks: It tracks events that may impact the robot's future navigation tasks, such as elevator maintenance notifications. This is done once at the beginning of a navigation task. Additionally, this module validates paths generated by the \textit{osmAGPathPlanner} against tracked events (closed areas). 
	
	When information about an external event is received, the LLM determines whether it affects the current or future robot navigation task. If so, the LLM generates a JSON string, which is a human language format summary of this event as well as all previously saved events, which the module stores as event tracking data for later use during path approval. See Fig. \ref{fig:flow} for an example. In addition to notifications, the module also tracks general information, such as time, it removes events from tracking once they are no longer relevant.
	
	Additionally, this module validates paths generated by the \textit{osmAGPathPlanner} (Section. \ref{osmAG Path Planner}) by checking the event tracking data, ensuring the shortest path avoids invalid areas. 
	The module also accounts for the robot's capabilities. For instance, in our multi-floor experiment (Section. \ref{result_Multifloor_Navigation}), the module recognize that a wheeled robot should use elevators instead of stairs when navigating between floors.
	
	The prompt used in this module to assess the path is shown in Fig. \ref{fig:prompt}. It describes the task and specifies the input and output formats, without examples or chain-of-thought reasoning.
	The output of this module includes validation of the shortest path and, if the path is invalid, identifies which areas should be avoided and puts those in the `areas to avoid' field, denoted as $A_{\text{invalid}}$. Since the LLM can infer potential impacts, for instance, if a notification states that main corridors will be closed for classroom renovations, the LLM may infer that other corridors in the same sector might also be affected. An additional field called `areas try to avoid' denoted as $A_{soft}$, is included, assigning extra costs $d_{soft}$ to try to avoid these areas without avoiding them completely.
	{\color{black}Compared to pure A*, this module brings in additional computation time due to LLM inference, each online query taking on average about 2 seconds. However, by identifying and avoiding invalid or unfavorable areas, it ultimately reduces overall navigation time that would otherwise be spent traversing unsuitable paths.}
	
	\subsection{osmAG Path Planner}
	\label{osmAG Path Planner}
	
	This module utilizes the precomputed costs between passage pairs $d_{ij}$, incorporates additional passage costs $pc_{i}$, and integrates updates from the \textit{NavigationEventMonitor} if the previously planned path is determined invalid or the intermediate areas of $\mathbf{P}$ are in  $A_{soft}$.
	The total cost $t_{ij}$ of an edge \( (v_i, v_j) \) in $AG$ is given by:
	$$
	\vspace{-1mm}
	t_{ij} = d_{ij} + d_{soft}+pc_{i} + pc_{j}
	\vspace{-1mm}
	$$
	
	The passages associated with an inaccessible area \( A_{\text{invalid}} \) are denoted as \( E_{\text{invalid}} \subset E \). Removing them from \( E \) gives:
	$$\vspace{-1mm}
	E' = E \setminus E_{\text{invalid}}
	\vspace{-1mm}$$
	
	To incorporate arbitrary start and goal coordinates into the graph, we need to add the appropriate vertices and edges. For the robot's starting position \( P_{\text{robot}} \), we add edges between \( P_{\text{robot}} \) and all passages in the start area (\( A_{\text{start}} \)). Similarly, for the final destination, we include the centroid of the destination area and create edges connecting it to all passages within the destination area (\( A_{\text{goal}} \)). Let \( v_s \) represent the robot's starting position \( P_{\text{robot}} \), and \( v_d \) represent the centroid of the destination area.
	
	The updated set of vertices \( V' \) and edges \( E'' \) can be expressed as:
	$$
	\vspace{-1mm}
	V' = V \cup \{ v_s, v_d \} 
	\vspace{-0.2mm}$$
	$$
	\vspace{-0.2mm}
	E'' = E' \cup \{ (v_s, v_i) \mid v_i \in A_{\text{start}} \} \cup \{ (v_d, v_j) \mid v_j \in A_{\text{goal}} \}
	\vspace{-0.5mm}$$
	The final graph incorporating the start position and the destination centroid for this specific navigation task is:
	$$\vspace{-2mm}
	AG' = (V', E'')
	\vspace{-0.0mm}$$
	\vspace{-0mm}
	Finally path \( \mathbf{P} \) is defined as:
	$\mathbf{P} = A^*(AG')$, where $A^*$ represents the well-known A star algorithm.
	Fig. \ref{fig:pipe_map} demonstrates our approach on an example: The $A^*$ method selects the path in (a) only utilizing pre-computed $d_{ij}$. However, with insights from the \textit{PassageCostEvaluator}, the path in (a) is assigned a higher cost, leading to the path in (b). The \textit{NavigationEventMonitor} then excludes the invalid area (yellow area), resulting in the final approved path in (c).
	
	{\color{black}Utilizing osmAG's room-level topology as the underlying graph structure, instead of the high-resolution occupancy grid, substantially decreases the computational burden of A* search and allows for more efficient plan and re-plan.}
	\subsection{Integration osmAGPathPlanner with ROS move\_base}
	\label{move_base}
	After the path $\mathbf{P}=( v_{1}, v_{2}, \ldots, v_{k})$ receives approval from the \textit{NavigationEventMonitor} module, $( v_{1}, v_{2}, \ldots, v_{k})$ will be sent to to the move\_base actionlib \cite{actionlib} as goals. Concurrently, the osmAG map is rendered as an occupancy grid map and  published as the global map in move\_base. This is done by rendering the polygons as occupied cells in the free grid map and then painting the cells of chosen passages as free again. This rendered map is restricted to the chosen areas and the selected passages, ensuring that the move\_base global planner strictly adheres to the path provided by the \textit{osmAGPathPlanner} as illustrated in Fig. \ref{fig:pipe_map}(d). In move\_base, we use the commonly employed 'Navfn' as the global planner with the rendererd map and 'DWA' (Dynamic Window Approach) as the local planner with a local map created from the global grid map updated with LiDAR scans. Despite utilizing documented experiences from previous tasks and external notifications, the robot may still encounter inaccessible passages. In such cases, our method receives a failure message from move\_base as the rendered map is restricted to the selected areas and passages, preventing move\_base from finding an alternative path.
	When a passage failure occurs, it is documented, and the \textit{osmAGPathPlanner} module re-plans the path from current robot postion to $A_{goal}$ until the robot successfully reaches $A_{goal}$.

	\subsection{Documenting Passages Accessibility} 
	\label{sec:experience}
	During navigation, this module continuously documents the success or failure of each passage traversal.
	The $PassageCostEvaluator$ module sends this stored experience passage data to LLM to evaluate each passage cost $pc_{i}$.
	Furthermore, if a passage is found to be inaccessible during a trial in experiment, the module marks it as `infeasible' and excludes it from the $AG'$ during re-planning.

	\section{Experiments}
	The goal of this study is to integrate laser-based robot navigation using osmAG with external information. A direct comparison with other LLM-aided navigation methods is impractical due to several key differences. First, our approach relies exclusively on laser sensors, while most existing methods depend on camera-based inputs.  
	Second, our test environment is significantly larger, focusing on spaces where announcements and notifications are common—unlike smaller, household-scale environments like apartments, where they are rarely used.
	Additionally, our method integrates real external notifications, which would not make sense in other environments or systems. 
	These fundamental differences in sensor setup and environment scale limit the feasibility of a fair comparison with other existing LLM aided navigation.
	
	Therefore, we conducted our experiments in Gazebo simulation using an actual osmAG map of our campus building, enriched with real notifications from our campus administrative department. This setup allows us to evaluate our system's performance and compare it with the ROS move\_base framework. 
	In Section \ref{Experiments Setup}, we outline our experiment setup. Section \ref{sec:NavigationEventMonitor} details how we assess the LLMs' ability to track navigation-related information and validate a proposed shortest path. Section \ref{Comprehensive_System_Evaluation} compares our method with ROS move\_base and Section \ref{Ablation_Experiment} evaluates the contributions of the two LLM modules to our approach in an ablation study. Finally, Section \ref{experiment_multi_floor} demonstrates our system can operate across floors and avoid elevators under maintenance.
	
	\subsection{Experiments Setup}
	\label{Experiments Setup}
	\subsubsection{Robot and World}
	In our experiments, we simulate the `turtlebot3 waffle' robot equipped with a 6-meter LiDAR sensor. To ensure a fair comparison, the baseline move\_base package shares identical parameters with our approach. We use ROS Gazebo to simulate an environment based on an osmAG of a campus building, covering over 6,200 square meters and 70+ rooms per floor. 
	The baseline move\_base is utilizing a rendered osmAG of all areas and passages not occupied. This occupancy grid map is the global map, which is then updated using the LiDAR data. 
	
	In our experimental setup, we have confined the robot's operations primarily to the first floor for two main reasons. First, the move\_base package, the baseline, lacks multi-floor navigation ability. Second, allowing the robot access to multiple floors could simplify the avoidance strategy by merely shifting its route vertically (moving upstairs or downstairs) to bypass obstructed areas. By confining the tests to the first floor, we can better challenge the algorithm to navigate longer detours outside the building. 
	\begin{figure*}[t]
		\centering
		\includegraphics[width=0.90\textwidth]{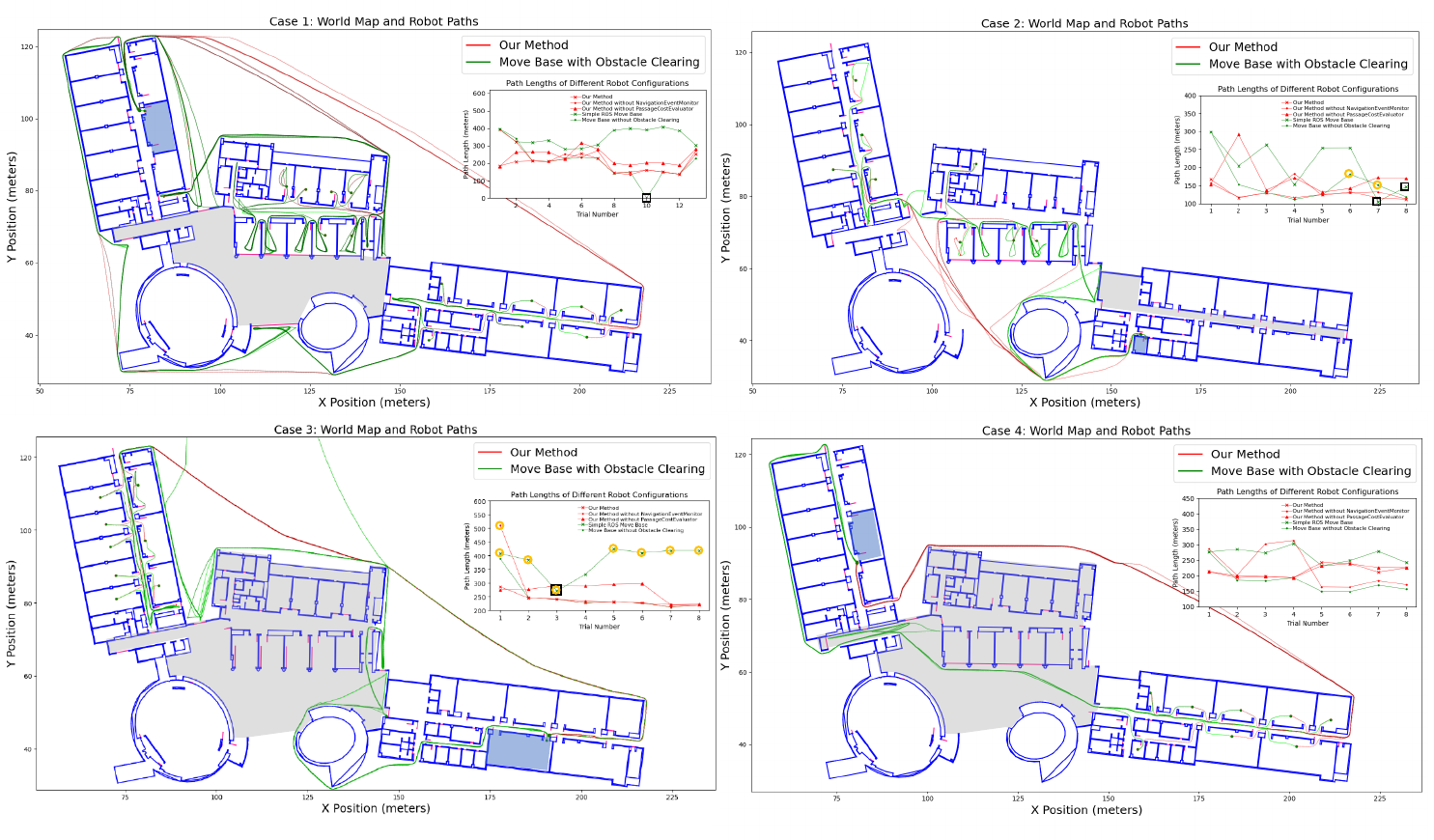}
		\caption{This figure presents the paths from four experimental cases. For clarity, only the paths from our method {\color{black}(red)} and move\_base with clearing between trials {\color{black}(green)} are displayed on the map. In the map, blue lines represent areas in the osmAG, while red lines indicate passages between them. Each case was tested in a Gazebo world with different door configurations (open or closed). 
			Grey shaded zones represent restricted areas caused by events in case 1): a graduation party in the lobby, 2) classroom renovation in the D sector, 3) a fire drill in the lobby and B sector, and 4) a wireless access upgrade in the lobby and B sector. Blue shaded zones represent destination areas of each case.
			The inset figures show the path lengths for each trial, comparing our method, our method with ablation testing, move\_base with and without obstacle layer clearing. The black box highlights where move\_base failed, and the data within the yellow circle shows where the path intruded into restricted areas. In Case 4, only the path with \textit{NavigationEventMonitor} avoids intruding into restricted areas.
		}
		\label{fig:case_path}
		\vspace{-7mm}
	\end{figure*}
	
	\subsubsection{Cases}
	To evaluate the system's ability to interpret external information, we designed 5 experimental cases (4 on the first floor, 1 spanning 2 floors), each involving a simulated instruction context and a real event notification. For example, an instruction might be: ``Hi, robot, please bring these tools back to the same room but on second floor.". This is then combined with event details sent to students by the campus administration, including specific times, events, durations, and locations, as illustrated in Fig. \ref{fig:flow}. 
	To test our approach to encode experience, in some cases, in the simulator, we close the doors of restricted areas, while in others, we leave them open. Those closed doors cause the according passage to be obstacles in the LiDAR map. The baseline move\_base will update the global map accordingly and plan then alternative path. For our method, our move\_base will report a failure to reach the goal point (next passage) and we mark that passage as closed and then re-plan the path as described in Section. \ref{move_base}.
	In each case, there are designated areas that the robot should consider restricted (shown in Fig. \ref{fig:case_path}), as indicated by prior notifications. Several starting areas are established for each case, along with destination areas, each pair forming one trial. The trials are designed such that the shortest path typically passes through restricted areas, requiring the robot, when notified, to intelligently plan a detour around them.
	
	Each case uses a specific Gazebo world, and the experience file in our method begins empty for each case. As the system navigates, each passage's accessibility is documented, gradually building a comprehensive understanding of the environment to optimize performance in subsequent trials.
	
	We defined 4 different cases of area closures through external information and then sampled 37 trials (start and goal area pairs). We then conducted test with our system (1x ours + 2x ablation study) and 2 versions of the baseline, 
	totaling 37 x 5 = 185 + 1 (multi-floor)
	navigation runs in Gazebo, which are explained in detail below. 
	
	%

	\subsection{NavigationEventMonitor LLM Experiments}
	\label{sec:NavigationEventMonitor}
	
	Here we test the first task of the \textit{NavigationEventMonitor} module, which assesses the relevance of each notification. We check the pure LLM performance of correctly identifying if a notification is or may become relevant. To evaluate the module’s ability to filter relevant information, we collected 10 navigation-related notifications and 10 unrelated ones (e.g., water cut-off notifications) and evaluated their correctness by hand. Additionally, after the module successfully tracks the events, we send notifications with a current time that is after the event’s end time. This tests the system’s ability to disregard outdated information. Success is defined as the module correctly identifying relevant notifications and removing event summaries once the event time has passed. 
	
	Furthermore, we evaluate the \textit{NavigationEventMonitor} module’s second task: identifying areas to avoid for affected \textit{osmAGPathPlanner} paths and confirming the validity of unaffected paths, based on the string of event summaries that was generated above. We conducted 37 trials across 4 cases, sending a total of 95 path approval requests to ChatGPT-4o, so we re-planned 58 times upon encountering a closed door. In this experiment we also report recall (the proportion of invalid paths correctly identified) and false positive rate (the proportion of valid paths incorrectly flagged as invalid). Alongside ChatGPT-4o, we wanted to also test DeepSeek-V3. However, due to the official API being unstable at the time of writing, evaluation was limited to 40 path approval requests, half of which were valid and the other ones invalid. Thus, for navigation simulations, we only used ChatGPT-4o.

	\subsection{Comprehensive Comparison to Baseline}
	\label{Comprehensive_System_Evaluation}
	With the above mentioned 37 trials in 4 cases we comprehensively test the whole navigation system and compare our method to the ROS move\_base baseline, with and without clearing obstacle layer of the global map of move\_base between trials. Note that the results from move\_base without obstacle layer clearing closely approximated the actual shortest path in later trials in one case. But in real life implementation, the global map needs periodic clearing to prevent sensor noise, environmental changes, or dynamic objects from being treated as permanent.
	The performance is evaluated based on the total navigation path length and whether the robot enters restricted areas.

	\subsection{Ablation Experiment}
	\label{Ablation_Experiment}
	\vspace{-1mm}
	To evaluate the contribution of individual components in our system, we conducted an ablation study on two key modules: the \textit{PassageCostEvaluator} and the \textit{NavigationEventMonitor}.
	In the \textit{PassageCostEvaluator} ablation, we removed this module, leaving only the \textit{NavigationEventMonitor} to track events, allowing the system to avoid areas without relying on past experience.
	In the \textit{NavigationEventMonitor} ablation, we excluded this module, enabling the system to use historical data but without current event tracking, to assess its importance in overall performance.
	
	\vspace{-1mm}
	\subsection{Multi-floor}
	\label{experiment_multi_floor}
	\vspace{-0mm}
	We also tested our system's ability to plan across floors, specifically avoiding obstacles like an elevator under maintenance. This test challenged the system to account for the robot’s limitations, such as avoiding the broken elevator and stairs, which are inaccessible for a wheeled robot, while still computing the shortest feasible path. 
	Although a robot’s physical capabilities remain constant during navigation and could theoretically be hard-coded, we argue that it is essential for the LLM to understand the robot's capabilities in order to make dynamic decisions, such as determining whether the robot can safely traverse small steps or slopes.
	
	\vspace{-0.5mm}
	\section{Results}
	\vspace{-0.5mm}

	\subsection{Effectiveness of NavigationEventMonitor}
	
	The \textit{NavigationEventMonitor} demonstrated high accuracy in assessing the relevance of notifications to the robot's navigation tasks. 
	Out of 20 notifications (10 navigation relevant, 10 irrelevant), the LLM correctly identified whether each notification was relevant to robot navigation. Additionally, the model successfully identified and resolved all 20 outdated notifications by deleting out of date events.
	
	The path validation results are summarized in Table \ref{table:deepseek} and evaluated by hand in terms of accuracy, recall, and false positive rate. Across 37 trials, 95 path approval requests were sent to ChatGPT-4o: 57 of 61 valid paths were approved (with is\_valid set to 'true' or the areas to avoid not in the path), and all 34 invalid paths were correctly flagged by setting is\_valid to false. However, 14 requests out of 95, the responses were semantically correct but deviated from the specified output format, such as returning `B sector' or a room number instead of the exact area name, such that our navigation module could not parse them correctly.
	For DeepSeek-V3, 40 path approval requests were sent, evenly split between valid and invalid paths. All responses adhered strictly to the required output format.
	The findings highlight the high performance of both models in path assessment, confirming their practical utility in real-world applications.

	\begin{table}[t]
		
		\vspace{-0mm}
		\caption{{Path Validation Comparison: ChatGPT-4o vs. DeepSeek-V3 (\%) }}
		\label{table:deepseek}		
		\centering
		\setlength{\tabcolsep}{1.2mm}
		\renewcommand{\arraystretch}{0.90}
		{
			\scalebox{0.90}{
				\begin{tabular}{ccc}
					\toprule
					Model&Accuracy &Recall
					\\
					\midrule
					ChatGPT-4o &0.96&1.0  \\
					DeepSeek-V3 
					&0.93&0.9\\
					\bottomrule
				\end{tabular}
		}}
		\vspace{-7mm}
	\end{table}
	
	\begin{table}[th]
		
		\vspace{-0mm}
		\caption{{Comparison of Average Path Lengths (m) Across Different Navigation Configurations for Various Cases }}
		\label{table:path_length}		
		\centering
		\setlength{\tabcolsep}{1.2mm}
		\renewcommand{\arraystretch}{0.90}
		{
			\scalebox{0.95}{
				\begin{tabular}{ccccc}
					\toprule
					Navigation configurations&Case 1 &Case 2&Case 3&Case 4
					\\
					\midrule
					Our method &\textbf{192.5}&\textbf{126.5}& \textbf{236.3}&214.7  \\
					Our method w/o 
					\textit{NavigationEventMonitor}  &221.2&135.4& 264.3& 223.3\\
					Our method w/o  \textit{PassageCostEvaluator} & 234.6& 171.4& 270.0& 215.9  \\
					Move\_base & 347.3& 202.2& 398.5& 267.5  \\
					Move\_base w/o  obstacle layer clearing & 220.1& 158.7& 248.1& \textbf{183.1}  \\

					\bottomrule
				\end{tabular}
		}}
		\vspace{-7mm}
	\end{table}

	\subsection{Comprehensive Comparison to Baseline Results}
	The results highlight significant differences in navigation performance between our method and baseline move\_base. Our system, utilizing osmAG based navigation with external notifications and former experience, demonstrated superior ability to avoid restricted areas and adapt to the environment. 
	In contrast, baseline move\_base performance varied depending on the obstacle layer's status. When the obstacle layer was cleared between trials, move\_base often chose inefficient paths through closed doors and then re-planning. Without clearing, move\_base improved in later trials, approximating the shortest path more closely. Regarding the interpretation of external information, move\_base is of course performing very badly, because it does not have such capability, while our approach worked very well.
	
	As shown in the inset figures in Fig. \ref{fig:case_path}, our method outperformed move\_base in the first trial of all four cases due to external information on restricted areas. In subsequent trials, our method, with stored experience, performed comparably to baseline move\_base without obstacle layer clearing. Overall, our approach consistently outperformed move\_base by maintaining context-aware path planning, demonstrating the effectiveness of integrating external information through \textit{PassageCostEvaluator} and \textit{NavigationEventMonitor}. Across 37 trials in 4 cases, our system consistently avoided restricted areas and successfully reached the designated destinations, reducing the total path length by 58\% compared to baseline move\_base with obstacle layer clearing between trials, as shown in Table \ref{table:path_length}. Notably, in cases 2, 3, and 4 (Fig. \ref{fig:case_path} (b), (c), (d)), where not all the doors to restricted areas were closed, the baseline move\_base entered restricted areas 25 times and failed 4 times across 74 runs in 37 trials, highlighting the clear advantage of our approach.

	\subsection{Ablation Experiment}
	As illustrated in Fig. \ref{fig:case_path} and Table \ref{table:path_length}, the ablation studies provided insight into the contribution of individual components to the system’s overall performance.
	\subsubsection{Ablation of PassageCostEvaluator}
	When the \textit{PassageCostEvaluator} is removed, the system still managed to avoid restricted areas but kept trying to pass infeasible passages over trials, leading to longer detours. This highlighted the importance of the \textit{PassageCostEvaluator} in leveraging historical data to optimize navigation decisions.
	\subsubsection{Ablation of NavigationEventMonitor}
	
	Removing the \textit{NavigationEventMonitor} significantly reduced the system's ability to avoid restricted areas. 
	In test cases where doors were closed in restricted areas, the robot eventually collected enough experience to perform similarly to our full method, though the initial few trials resulted in longer paths. However, in tests where doors remained open in restricted areas, the system without this module violated the restrictions introduced by the notifications. This experiment demonstrate the importance of the \textit{NavigationEventMonitor} in ensuring compliance with external information during navigation.
	
	\subsection{Multi-floor Navigation}
	\vspace{-0mm}
	\label{result_Multifloor_Navigation}
	
	In the multi-floor navigation test, our system effectively planned a path across different floors, particularly by avoiding areas like elevators under maintenance and stairs that the wheeled robot cannot climb. The \textit{osmAGPathPlanner} proposed a path through the stairs, but the \textit{NavigationEventMonitor} rejected it, based on our prompt that this is a wheeled robot and by looking up in the osmAG map, that this particular area is a stair. The accompanying video illustrates this capability, highlighting a distinct advantage over move\_base, which lacks support for multi-floor navigation and does not consider the robot's capabilities. 
	
	\section{Conclusion and Limitation}
	\vspace{-1mm}
	This paper presents the integration of Large Language Models into a robot path planning and navigation system, combining the efficiency and speed of the $A^*$ algorithm with the broad knowledge and reasoning capabilities of LLMs. This hybrid approach leverages the strengths of both methods: $A^*$ for precise, optimized pathfinding, and LLMs for understanding and adapting to human instructions and societal norms, utilizing our osmAG map representation, which is very suitable for LLMs.
	
	Although this study was conducted using a map limited to our campus building and its associated notifications, we have demonstrated that current LLMs, possess the capability to comprehend textual maps and integrate real-world societal information. 
	In the future, as the system is tested in more diverse real-world environments, we aim to collect and incorporate more detailed human preferences and contextual data. This could include avoiding crowded areas during peak hours or steering clear of sensitive locations like VIP conference rooms.
	This study lays a promising foundation for the continued integration of LLMs with traditional algorithms into everyday applications, paving the way for more intelligent and socially adaptive robotic systems in the future.
	\vspace{-1mm}
	
	
	
	\bibliographystyle{IEEEtran}
	\bibliography{Bibliography}
	
\end{document}